\pdfoutput=1

\documentclass[11pt]{article}

\usepackage[]{acl}  
\usepackage{nert}

\usepackage{times}
\usepackage{latexsym}

\usepackage{graphicx} 
\usepackage{caption}
\usepackage{amsmath}
\RestyleAlgo{ruled}
\LinesNumbered
\newcommand{\Lagr}{\mathcal{L}}
\newcommand{\fancyS}{\mathcal{S}}

\usepackage[utf8]{inputenc}
\usepackage{todonotes}
 \usepackage{multirow}
 \usepackage[table,xcdraw]{xcolor}

\title{Adversarial Lens: Exploiting Attention Layers to Generate Adversarial Examples for Evaluation\\[15pt]}

\author{Kaustubh Dhole \\
  Department of Computer Science \\
  Emory University \\
  \eml{kdhole@emory.edu}
}

\begin{document}
\maketitle
\begin{abstract}
Recent advances in mechanistic interpretability suggest that intermediate attention layers encode token-level hypotheses that are iteratively refined toward the final output. In this work, we exploit this property to generate adversarial examples directly from attention-layer token distributions. Unlike prompt-based or gradient-based attacks, our approach leverages model-internal token predictions, producing perturbations that are both plausible and internally consistent with the model’s own generation process. We evaluate whether tokens extracted from intermediate layers can serve as effective adversarial perturbations for downstream evaluation tasks. We conduct experiments on argument quality assessment using the ArgQuality dataset, with LLaMA-3.1-Instruct-8B serving as both the generator and evaluator. Our results show that attention-based adversarial examples lead to measurable drops in evaluation performance while remaining semantically similar to the original inputs. However, we also observe that substitutions drawn from certain layers and token positions can introduce grammatical degradation, limiting their practical effectiveness. Overall, our findings highlight both the promise and current limitations of using intermediate-layer representations as a principled source of adversarial examples for stress-testing LLM-based evaluation pipelines.
\end{abstract}

\section{Introduction}
Recent efforts in mechanistic interpretability have highlighted the wealth of information encoded within the layers of large language models (LLMs)~\cite{meng2022locating,sharkey2025open}. These layers, which are often overlooked in favor of the final outputs, have been shown to act as iterative predictors of the eventual response~\cite{jastrzebski2018residual,lesswrong_logitlens}, providing insights into the model’s generation process. 
While most of these techniques have heavily focused on interpretability, we argue that they could potentially be adapted to generate paraphrastic and adversarial examples for evaluation tasks -- which generally operate over model generated data.

Probing the attention layers has multiple advantages -- First, since the layers act as both iterative indicators of the final output~\cite{belrose2023eliciting}, and store related entities~\cite{meng2022locating,hernandezlinearity}, they can provide natural language variations by potentially treating LLMs as knowledge bases. Second, these generations, can be obtained early on without necessitating running over all the layers~\cite{din2024jump,pal2023future}. Third, these generations might provide cues for model hallucinations~\cite{yuksekgonul2024attention} as gradual deviations from the original token are obtained from the model itself. 
From an adversarial point of view, tokens from intermediate layers are valuable as they can act as perturbations to the original input. Specifically, the outputs are~\textit{iteratively refined}, since as the activations move towards the last layer they tend to move towards the direction of the negative gradient~\cite{jastrzebski2018residual} or each successive layer achieving lower perplexity~\cite{belrose2023eliciting}. This is also precisely how adversarial examples are constructed -- perturbing towards the direction of the positive gradient of the loss~\cite{fgsm}.

Hence, in this study, we explore whether such fine-grained information extracted from the attention layers of large language models (LLMs) can be leveraged to generate adversarial examples on downstream natural language tasks, particularly critical tasks such as evaluation.

Specifically, in this work, we introduce two attention-based adversarial generation methods: attention-based token substitution and attention-based conditional generation, both of which leverage intermediate-layer token predictions to construct plausible yet adversarial inputs.

Our paper is organized as follows:~\cref{relwork} first discusses related work in interpretability and generation evaluation.~\cref{example_generation} and~\cref{methodsexp} defines and implements the two approaches for generating examples.~\cref{results} finally discusses the results and analysis. 

\section{Related Work}\label{relwork}
We now discuss some of the related work in mechanistic interpretability and LLM based evaluation to place our work in context. 

Recent interpretability studies have explored the information encoded within the internal layers of large language models (LLMs) to better understand how models generate subsequent tokens. For instance, LogitLens and TunedLens~\cite{lesswrong_logitlens, belrose2023eliciting} demonstrate that intermediate layers can be made to predict tokens similar to those generated in the final layer, by attaching a trained or untrained unembedding matrix to them. Methods such as ROME~\cite{meng2022locating} highlight the role of specific components, like MLP layers, in acting as key-value stores~\cite{geva-etal-2021-transformer} that retain critical entities related to the input, such as associating ``Seattle'' with ``Space Needle.'' Besides, LLMs assign greater attention to constraint tokens when their outputs are factual vis-à-vis when they are hallucinating~\cite{yuksekgonul2024attention}. These findings underscore the utility of the attention layers beyond interpretability but also as knowledge probes~\cite{alain2016understanding}, to extract inherent knowledge. Hence, in this work, we explore if the information from these layers can act as a resource for generating adversarial examples. 

Some techniques have been explored to probe intermediate layers to reveal token predictions~\cite{belrose2023eliciting} for interpretability and early exiting to improve inference time~\cite{geva-etal-2021-transformer}.

ResNets~\cite{jastrzebski2018residual} have been shown to perform iterative feature refinement (where
each block improves slightly but keeps the semantics of the representation of the previous layer)).

On the other hand, evaluation of LLM-generated outputs has become increasingly important as models are deployed in high-stakes settings. LLMs-as-judges have emerged as a common paradigm for evaluating generated text, either through prompting or via reward models trained on human preferences. In retrieval-augmented generation (RAG) systems, LLM-based evaluators are frequently used to assess dimensions such as answer quality, groundedness, and context relevance~\cite{dhole2025retrieve,Dhole_PyTerrier_Genrank}. Recent work has also begun to examine the robustness of such evaluators, demonstrating that groundedness and factuality judgments can be manipulated through adversarial perturbations~\cite{dhole-etal-2025-adversem}. These findings motivate a closer examination of whether adversarial examples derived from model internals pose additional risks to LLM-based evaluation frameworks.

\begin{figure}
    \centering
    \includegraphics[width=1\linewidth]{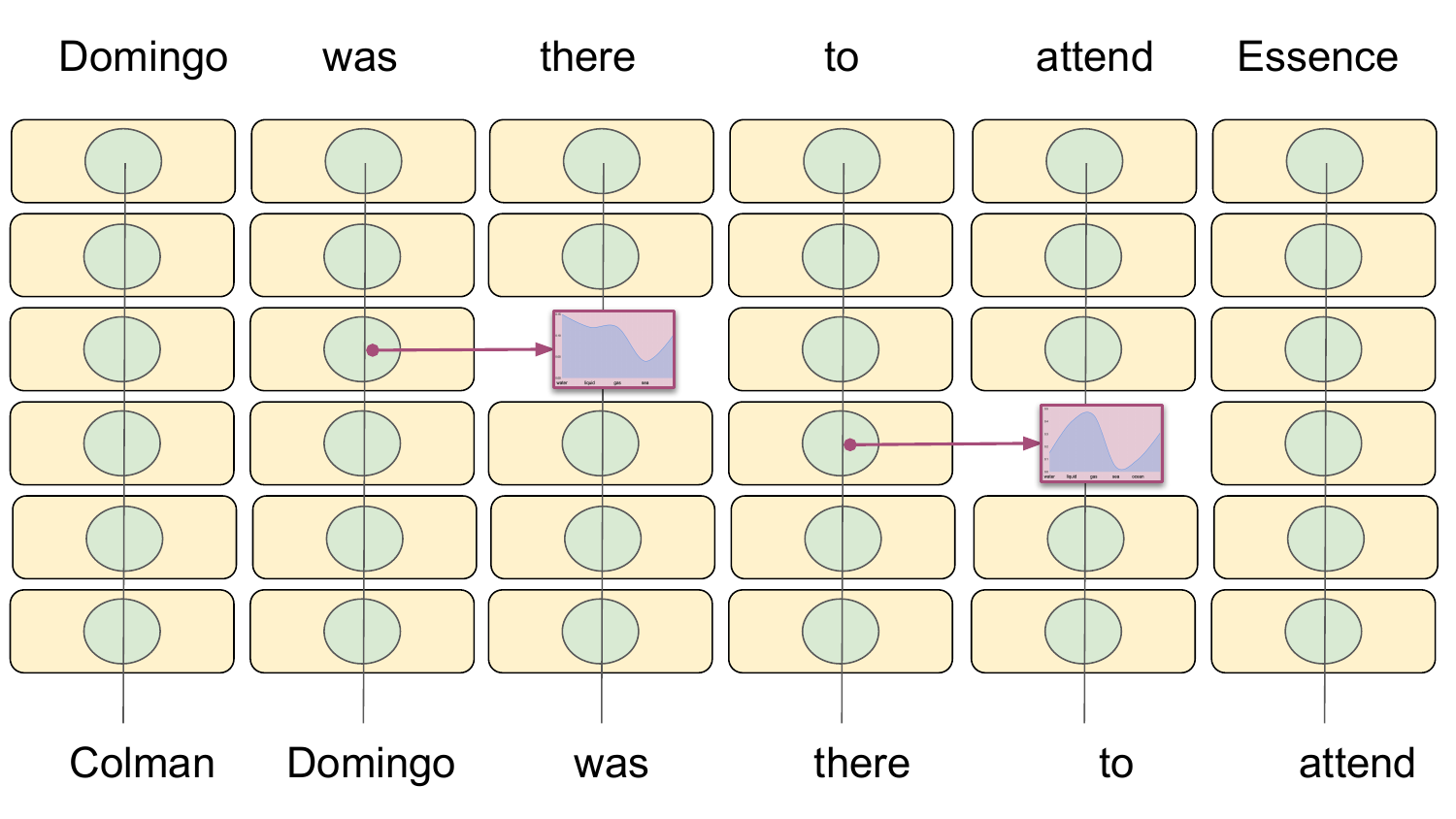}
    \caption{Tokens are extracted from intermediate layers to generate adversarial examples}
    \label{fig:layerwise}
\end{figure}

\section{Example Generation}\label{example_generation}
Prompt-based methods generate adversarial examples by instructing an LLM to rewrite or manipulate a given input according to a natural language directive~\cite{dhole2024conqret,dhole2024llmjudge}. Typically, the prompt includes a task description, an example instance, and its associated label, followed by an instruction to generate an adversarial or misleading variant.

While such methods are flexible and easy to apply in black-box settings, the generated examples may not correspond to naturally occurring model mistakes. As a result, prompt-based adversaries may diverge substantially from the original input distribution and fail to reflect the kinds of errors that arise organically during model generation. For this reason, we focus our study on attention-based methods that exploit the model’s own intermediate token predictions.

\begin{figure*}
    \centering
    \includegraphics[width=1\textwidth]{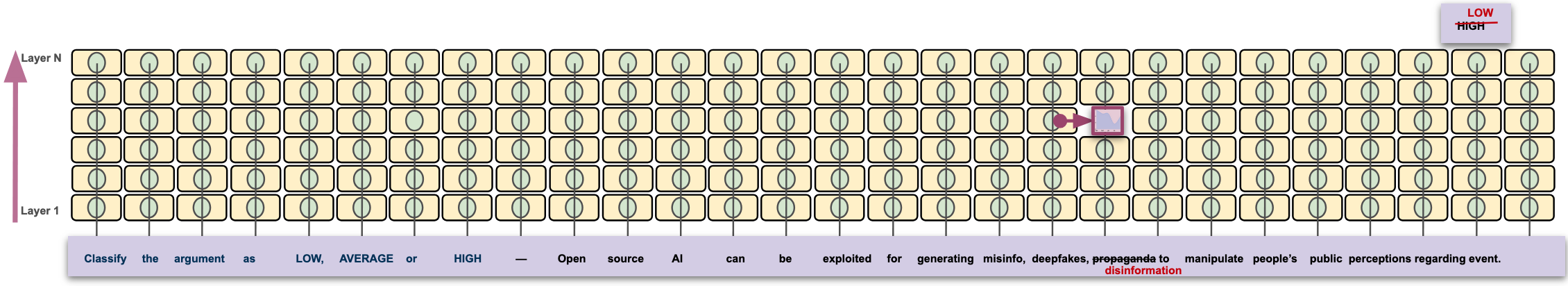}
    \caption{This is the expanded figure with the complete evaluation prompt. Here, the token `disinformation' is used to manipulate the model into changing the rating of the argument.}
    \label{fig:enter-label}
\end{figure*}
\subsection{Attention Based Methods}
In this section, we discuss our two methods of generating examples and how we use them for evaluation. Both the methods start by substituting a token $t_x$ from a given text sequence.

\textbf{Tokens Predicted by Attention Layers} In order to extract a novel token $t_x'$, we first gather the token distributions of all the layers at all the positions. To do the same, we pass the input sequence through a LLaMA-3.1 instruct 8B~\cite{llama-3.1} by first lens-tuning~\cite{belrose2023eliciting} it over the final layer's vocabulary. Lens-tuning, helps gauge the token distribution at each attention head and each layer as shown in~\cref{fig:layerwise}. Specifically, the token distribution is obtained by passing each attention layer's outputs through a linear layer and an unembedding matrix. Lens-tuning involves training this layer to minimize the KL-divergence between the vocabulary distribution of the final layer and that of an intermediate layer.\footnote{The tuned lens model is available on HuggingFace~\cite{wolf2020transformers} at~\url{hf.co/kdhole/Llama-3.1-8B-Instruct-tuned-lens}}

\subsection{Attention Token Substitution}
This approach involves substituting a single token from an input sequence with a novel token estimated from the attention layers' distribution. Formally, the transformation $T$ for a given $n$-token sequence $t_{1:n}$ perturbed at position $x \in [1,n]$ can be described as follows:
\[
T: (t_1, \ldots, t_{x-1}, t_x, t_{x+1}, \ldots) \rightarrow 
(t_1, \ldots, t_{x-1}, \textcolor{brown}{t_x'}, t_{x+1}, \ldots)
\]
where \( t_x' = A(l,x)\) is selected from the attention token distribution $A$ at position $x$ of an attention layer $l$, representing a plausible yet distinct alternative to the original token \( t_x \). 

This method is lightweight and is likely to introduce minor semantic changes, allowing for efficient token-level editing with minimal disruption to the overall meaning. 

\subsection{Attention Token Conditioned Generation}
We propose a second method to address the potential syntactic inconsistencies of the previous token substitution. In this approach, the substituted sequence until token $x$ serves as input for autoregressively generating the remaining sequence tokens, ensuring coherence and fluency. The transformation is described as:
\[
\begin{aligned}
T: (t_1, \ldots, t_x, t_{x+1}, \ldots, t_n) 
&\rightarrow
(t_1, \ldots, \textcolor{brown}{t_x', t_{x+1}', \ldots, t_m'})
\end{aligned}
\]

In this case, the model autoregressively generates all tokens after position $x$ viz., \( t_{x+1}', \ldots, t_n' \) by conditioning on the modified sequence \(t_1, \dots, t_{x-1}, \textcolor{brown}{t_x'} \), thereby aligning the entire sequence syntactically and semantically. The first token \( t_x' = A(l,x)\) is obtained as earlier from the attention distribution.

This method produces novel and diverse examples while ensuring syntactic correctness and coherence of the sequence. However, the newly generated tokens may introduce shifts in meaning, diverging from the original context. To minimize this effect, we choose token positions to substitute at the latter parts of the sequence so that there is a large overlap in the initial context used to dictate the rest of the tokens. The full procedure for adversarial example generation, including token and layer selection, is summarized in Algorithm~\ref{alg:token_substitution}.




\begin{algorithm}[ht]
\small
\caption{\small Adversarial Example Generation}
\label{alg:token_substitution}

\textbf{Inputs:} Input sequence $t_{1:n}$; model $M$; tuned-lens method $\Lagr$; token/layer selection method $\mathcal{S}$\;
\textbf{Output:} Modified sequence $t'_{1:n}$\;

\BlankLine
\textbf{Train tuned-lens model (done once):}\;
$M_\Lagr \gets \Lagr(M)$\;

\BlankLine
\textbf{Compute internal token distributions:}\;
$A = M_L(t_{1:n})$ where $A \in \mathbb{R}^{L \times n}$\;

\BlankLine
\textbf{Select token/layer position(s):}\;
$(x,l) = \fancyS(M_L, A)$\;
$t' = A(x,l)$\;

\BlankLine
\textbf{Substitute token:}\;
Replace $t_x$ with $t'$ to obtain $t'_{1:n}$\;

\BlankLine
\Return $t'_{1:n}$\;
\end{algorithm}

\section{Methods and Experiments}\label{methodsexp}

For evaluation, we use the LLaMA-3.1-Instruct-8B model~\cite{dubey2024llama}. We focus exclusively on argument quality assessment using the ArgQuality corpus~\cite{habernal2016argument}, a task well-suited for token-level perturbations since small lexical changes can significantly alter perceived argument strength.

ArgQuality classifies arguments into three categories: \emph{low}, \emph{average}, and \emph{high} quality. We construct evaluation instances in the form of (topic, stance, chosen argument, rejected argument) tuples and measure how often the model correctly prefers the higher-quality argument. This setting allows us to directly assess whether attention-based adversarial examples can degrade evaluation performance without substantially altering semantic content.

We first assess~\textbf{answer quality}, by evaluating whether these examples are useful for testing the robustness capabilities of both trained and fine-tuned models. This was done by first evaluating the model's performance on ArgQuality's test set and the modified test set generated from the two methods discussed in~\cref{example_generation}. Our evaluation set consists of 75 test examples from the ArgQuality corpus which we transform in the form of (topic, stance, chosen argument, rejected argument) tuples. 

\textbf{Adverserial Example Generation:} We lens-tune the LLaMA-3.1-Instruct-8B model so that we can display the tokens at each layer~\cite{belrose2023eliciting} and use them for generating adversarial sequences. We specifically generate the adversarial counterparts for the chosen and rejected arguments in the same. In our experiments, we choose $x=10$ and $l=18$. For evaluation, we use a few-shot prompt displayed in~\cref{fig:prompt_fn}.

\begin{table}[t]
\begin{tabular}{rrrrrrrllllllll}
\multicolumn{1}{l}{\textbf{L\textbackslash T}} & \textbf{16} & \textbf{32} & \textbf{48} & \textbf{64} & \textbf{80} & \textbf{128} &  &  &  &  &  &  &  &  \\
\textbf{28} & \cellcolor[HTML]{A7DAC2}.423 & \cellcolor[HTML]{8BCFAE}.403 & \cellcolor[HTML]{77C7A0}.389 & \cellcolor[HTML]{BBE1D0}.437 & \cellcolor[HTML]{F8F9FA}.479 & \cellcolor[HTML]{A4D8BF}.420 &  &  &  &  &  &  &  &  \\
\textbf{24} & \cellcolor[HTML]{93D2B4}.408 & \cellcolor[HTML]{8BCFAE}.403 & \cellcolor[HTML]{9FD6BC}.417 & \cellcolor[HTML]{A7DAC2}.423 & \cellcolor[HTML]{EDF4F2}.471 & \cellcolor[HTML]{77C7A0}.389 &  &  &  &  &  &  &  &  \\
\textbf{20} & \cellcolor[HTML]{CFE9DE}.451 & \cellcolor[HTML]{8BCFAE}.403 & \cellcolor[HTML]{57BB8A}.366 & \cellcolor[HTML]{CFE9DE}.451 & \cellcolor[HTML]{B0DDC7}.429 & \cellcolor[HTML]{8BCFAE}.403 &  &  &  &  &  &  &  &  \\
\textbf{16} & \cellcolor[HTML]{93D2B4}.408 & \cellcolor[HTML]{77C7A0}.389 & \cellcolor[HTML]{7FCAA6}.394 & \cellcolor[HTML]{7FCAA6}.394 & \cellcolor[HTML]{77C7A0}.389 & \cellcolor[HTML]{5EBD8F}.371 &  &  &  &  &  &  &  &  \\
\textbf{12} & \cellcolor[HTML]{93D2B4}.408 & \cellcolor[HTML]{63BF92}.375 & \cellcolor[HTML]{CFE9DE}.451 & \cellcolor[HTML]{8BCFAE}.403 & \cellcolor[HTML]{57BB8A}.366 & \cellcolor[HTML]{57BB8A}.366 &  &  &  &  &  &  &  &  \\
\textbf{8} & \cellcolor[HTML]{F8F9FA}.479 & \cellcolor[HTML]{9FD6BC}.417 & \cellcolor[HTML]{77C7A0}.389 & \cellcolor[HTML]{7FCAA6}.394 & \cellcolor[HTML]{77C7A0}.389 & \cellcolor[HTML]{77C7A0}.389 &  &  &  &  &  &  &  & 
\end{tabular}
\caption{Effect of Attention Token Substitution using adversarial tokens from different layers (L) and different token positions (T). }
\label{tab:layerwise_and_token_position_wise_results1}
\end{table}

\begin{table}[t]
\begin{tabular}{rrrrrrrllllllll}
\multicolumn{1}{l}{\textbf{L\textbackslash T}} & \textbf{16} & \textbf{32} & \textbf{48} & \textbf{64} & \textbf{80} & \textbf{128} &  &  &  &  &  &  &  &  \\
\textbf{28} & \cellcolor[HTML]{65C093}.280 & \cellcolor[HTML]{9DD6BB}.400 & \cellcolor[HTML]{C3E4D5}.480 & \cellcolor[HTML]{9AD4B8}.393 & \cellcolor[HTML]{7ECAA5}.333 & \cellcolor[HTML]{76C69F}.316 &  &  &  &  &  &  &  &  \\
\textbf{24} & \cellcolor[HTML]{7CC9A4}.330 & \cellcolor[HTML]{9DD6BB}.400 & \cellcolor[HTML]{8FD0B1}.370 & \cellcolor[HTML]{84CCA9}.346 & \cellcolor[HTML]{9DD6BB}.400 & \cellcolor[HTML]{CDE8DC}.500 &  &  &  &  &  &  &  &  \\
\textbf{20} & \cellcolor[HTML]{6EC49A}.300 & \cellcolor[HTML]{94D2B4}.380 & \cellcolor[HTML]{ACDBC5}.430 & \cellcolor[HTML]{7BC9A3}.328 & \cellcolor[HTML]{5DBD8E}.263 & \cellcolor[HTML]{CDE8DC}.500 &  &  &  &  &  &  &  &  \\
\textbf{16} & \cellcolor[HTML]{78C7A0}.320 & \cellcolor[HTML]{A2D8BE}.410 & \cellcolor[HTML]{8ACFAE}.360 & \cellcolor[HTML]{B0DDC7}.439 & \cellcolor[HTML]{BFE3D2}.471 & \cellcolor[HTML]{F8F9FA}.591 &  &  &  &  &  &  &  &  \\
\textbf{12} & \cellcolor[HTML]{65C093}.280 & \cellcolor[HTML]{BAE1CE}.460 & \cellcolor[HTML]{94D2B4}.380 & \cellcolor[HTML]{C4E5D6}.482 & \cellcolor[HTML]{D9EDE4}.526 & \cellcolor[HTML]{B5DFCB}.450 &  &  &  &  &  &  &  &  \\
\textbf{8} & \cellcolor[HTML]{57BB8A}.250 & \cellcolor[HTML]{A2D8BE}.410 & \cellcolor[HTML]{9DD6BB}.400 & \cellcolor[HTML]{A6D9C1}.418 & \cellcolor[HTML]{86CDAA}.350 & \cellcolor[HTML]{61BF91}.273 &  &  &  &  &  &  &  & 
\end{tabular}
\caption{Effect of Attention Token Conditioned
Generation using adversarial tokens from different layers (L) and different token positions (T). }
\label{tab:layerwise_and_token_position_wise_results2}
\end{table}

\begin{figure*}[ht]
\centering
\small
\begin{tabular}{|>{\ttfamily\raggedright\arraybackslash}p{0.95\textwidth}|}
\hline

\hspace*{1em}"Rate the quality of the given argument among 'low', 'average' and 'high'. Just mention either of the options and do not provide an explanation. The argument should be rated high if it convinces the reader towards the expected stance for a controversial topic.\textbackslash n"\newline
\hspace*{1em}"\textbackslash n\textbackslash nThe topic is 'Ban Plastic Water Bottles'.\textbackslash nThe stance is 'No bad for the economy'.\textbackslash nHere is the argument: U.S. alone grew by over 13\%. According to research and consulting done by the Beverage Marketing Corporation, the global bottled water industry has exploded to over \$35 billion. Americans alone paid \$7.7 billion for bottled water in 2002. In 2001, for example, globally bottled water companies produced over 130,000 million liters of water. This produced roughly 35,000 million dollars in revenue for the world's thousands of bottled water companies in 2001.\textbackslash nRating: 'average'"\newline
\hspace*{1em}"\textbackslash n\textbackslash nThe topic is 'Is porn wrong'.\textbackslash nThe stance is 'Yes porn is wrong'.\textbackslash nHere is the argument: Porn is definitely wrong. Porn is like an addiction to some people which is unhealthy and can lead to guilt and lust. An addiction to porn gives an unhealthy image of real sex. Porn promotes the fact that sex is totally based on pleasure, but it is actually based on love and affection also. Porn inspired numerous crimes that sometimes abuse the rights and virginity of many people.\textbackslash nRating: 'high'"\newline
\hspace*{1em}"\textbackslash n\textbackslash nThe topic is 'William Farquhar ought to be honoured as the rightful founder of Singapore'.\textbackslash nThe stance is 'Yes of course'.\textbackslash nHere is the argument: Farquhar contributed significantly, even forking out his own money to start up the colony carved out of the jungle, by first offering money as an incentive for people to hunt and to exterminate rats and centipedes. Raffles did nothing of that sort.\textbackslash nRating: 'low'"\newline
\hspace*{1em}f"\textbackslash n\textbackslash nThe topic is \{\textbf{topic}\}.\textbackslash nThe stance is \{\textbf{stance}\}.\textbackslash nHere is the argument: \{\textbf{arg}\}\textbackslash nRating:"\newline
\\
\hline
\end{tabular}
\caption{The few-shot prompt used for rating argument quality.}
\label{fig:prompt_fn}
\end{figure*}
\section{Results}\label{results}
The results indicate that adversarial tokens extracted from a wide range of layers and token positions can negatively impact evaluation accuracy. However, we also observe that certain configurations—particularly substitutions at later token positions—can paradoxically improve performance. This suggests that not all intermediate-layer tokens act as effective adversarial perturbations, and that both layer depth and token position play a critical role in determining adversarial effectiveness. Tables~\ref{tab:layerwise_and_token_position_wise_results1} and~\ref{tab:layerwise_and_token_position_wise_results2} show that substitutions drawn from mid-to-late layers (e.g., layers 16–28) generally induce larger performance drops than those from earlier layers, particularly when applied at moderate token positions. However, very late token substitutions occasionally improve accuracy, likely by introducing clarifying or corrective lexical choices.

Table~\ref{tab:llm_preference} further confirms this trend at the aggregate level: evaluation accuracy drops from 0.42 to 0.34 in the few-shot setting and from 0.60 to 0.57 in the fine-tuned setting when adversarial examples generated via our attention-based methods are introduced. Together, these results indicate that attention-layer-derived perturbations can reliably degrade evaluator performance, though their impact is highly sensitive to where in the sequence and from which layer the token is extracted.

\begin{table}[t]
\centering
\resizebox{\columnwidth}{!}{%
    \begin{tabular}{lcc}
    \hline
                         & Few-shot       & Fine-tuned     \\
    \hline
    Original Test Set    & .42         & .60           \\
    Adversarial Lens Test Set & .34         & .57           \\
    \hline
    \end{tabular}
}
\caption{Argument Quality Evaluator Performance}
\label{tab:llm_preference}
\end{table}

\section{Conclusion}
Evaluation tasks provide a natural test bed for attention-based adversarial example generation, as LLM-based judges routinely consume model-generated text that may already contain subtle inconsistencies or errors. In this work, we show that intermediate attention layers can be exploited to generate adversarial examples without requiring access to the model’s final layer or gradients.

While our results demonstrate consistent successful attacks, we also find that many intermediate-layer substitutions lead to grammatical degradation, limiting their effectiveness as practical adversaries. In that regard, we introduce the attention token conditioned approach. Our preliminary study highlights the need for more selective token and layer selection mechanisms, and the potential to extract knowledge from intermediate layers gradually. 

Future work should explore principled criteria for identifying syntactically valid and semantically impactful substitutions, as well as extending these methods to domains where strict linguistic structure is less critical. For example, structured domains such as electronic health records, where substituting diagnosis or procedure codes may have significant downstream effects, present a promising direction for applying attention-based adversarial methods.

We focus on evaluation tasks rather than general classification, as evaluation models are especially likely to encounter generations influenced by near-final-layer representations. Overall, this study demonstrates that intermediate-layer representations offer a promising—but currently imperfect—source of adversarial examples for stress-testing LLM evaluation pipelines.
\section*{Acknowledgments}
The author thanks Eugene Agichtein from Emory University for insightful discussions.

\bibliography{mypaper}

\begin{thebibliography}{21}
\expandafter\ifx\csname natexlab\endcsname\relax\def\natexlab#1{#1}\fi

\bibitem[{Alain and Bengio(2016)}]{alain2016understanding}
Guillaume Alain and Yoshua Bengio. 2016.
\newblock \href {http://arxiv.org/abs/1610.01644} {Understanding intermediate layers using linear classifier probes}.

\bibitem[{Belrose et~al.(2023)Belrose, Furman, Smith, Halawi, Ostrovsky, McKinney, Biderman, and Steinhardt}]{belrose2023eliciting}
Nora Belrose, Zach Furman, Logan Smith, Danny Halawi, Igor Ostrovsky, Lev McKinney, Stella Biderman, and Jacob Steinhardt. 2023.
\newblock Eliciting latent predictions from transformers with the tuned lens.
\newblock \emph{arXiv preprint arXiv:2303.08112}.

\bibitem[{Dhole(2024)}]{Dhole_PyTerrier_Genrank}
Kaustubh Dhole. 2024.
\newblock \href {https://github.com/emory-irlab/pyterrier_genrank} {{PyTerrier-GenRank: The PyTerrier Plugin for Reranking with Large Language Models}}.

\bibitem[{Dhole and Agichtein(2024)}]{dhole2024llmjudge}
Kaustubh Dhole and Eugene Agichtein. 2024.
\newblock \href {https://github.com/emory-irlab/argumentation-rag/blob/main/LLM_Judges_for_Argumentation.pdf} {Llm judges for retrieval augmented argumentation}.

\bibitem[{Dhole et~al.(2025)Dhole, Chandradevan, and Agichtein}]{dhole-etal-2025-adversem}
Kaustubh Dhole, Ramraj Chandradevan, and Eugene Agichtein. 2025.
\newblock \href {https://doi.org/10.18653/v1/2025.starsem-1.32} {{A}dv{ERSEM}: Adversarial robustness testing and training of {LLM}-based groundedness evaluators via semantic structure manipulation}.
\newblock In \emph{Proceedings of the 14th Joint Conference on Lexical and Computational Semantics (*SEM 2025)}, pages 395--408, Suzhou, China. Association for Computational Linguistics.

\bibitem[{Dhole(2025)}]{dhole2025retrieve}
Kaustubh~D Dhole. 2025.
\newblock To retrieve or not to retrieve? uncertainty detection for dynamic retrieval augmented generation.
\newblock \emph{arXiv preprint arXiv:2501.09292}.

\bibitem[{Dhole et~al.(2024)Dhole, Shu, and Agichtein}]{dhole2024conqret}
Kaustubh~D. Dhole, Kai Shu, and Eugene Agichtein. 2024.
\newblock \href {http://arxiv.org/abs/2412.05206} {Conqret: Benchmarking fine-grained evaluation of retrieval augmented argumentation with llm judges}.

\bibitem[{Din et~al.(2024)Din, Karidi, Choshen, and Geva}]{din2024jump}
Alexander~Yom Din, Taelin Karidi, Leshem Choshen, and Mor Geva. 2024.
\newblock Jump to conclusions: Short-cutting transformers with linear transformations.
\newblock In \emph{Proceedings of the 2024 Joint International Conference on Computational Linguistics, Language Resources and Evaluation (LREC-COLING 2024)}, pages 9615--9625.

\bibitem[{Dubey et~al.(2024{\natexlab{a}})Dubey, Jauhri, Pandey, Kadian, Al-Dahle, Letman, Mathur, Schelten, Yang, Fan et~al.}]{llama-3.1}
Abhimanyu Dubey, Abhinav Jauhri, Abhinav Pandey, Abhishek Kadian, Ahmad Al-Dahle, Aiesha Letman, Akhil Mathur, Alan Schelten, Amy Yang, Angela Fan, et~al. 2024{\natexlab{a}}.
\newblock The llama 3 herd of models.

\bibitem[{Dubey et~al.(2024{\natexlab{b}})Dubey, Jauhri, Pandey, Kadian, Al-Dahle, Letman, Mathur, Schelten, Yang, Fan et~al.}]{dubey2024llama}
Abhimanyu Dubey, Abhinav Jauhri, Abhinav Pandey, Abhishek Kadian, Ahmad Al-Dahle, Aiesha Letman, Akhil Mathur, Alan Schelten, Amy Yang, Angela Fan, et~al. 2024{\natexlab{b}}.
\newblock The llama 3 herd of models.
\newblock \emph{arXiv preprint arXiv:2407.21783}.

\bibitem[{Geva et~al.(2021)Geva, Schuster, Berant, and Levy}]{geva-etal-2021-transformer}
Mor Geva, Roei Schuster, Jonathan Berant, and Omer Levy. 2021.
\newblock \href {https://doi.org/10.18653/v1/2021.emnlp-main.446} {Transformer feed-forward layers are key-value memories}.
\newblock In \emph{Proceedings of the 2021 Conference on Empirical Methods in Natural Language Processing}, pages 5484--5495, Online and Punta Cana, Dominican Republic. Association for Computational Linguistics.

\bibitem[{Goodfellow et~al.(2015)Goodfellow, Shlens, and Szegedy}]{fgsm}
Ian Goodfellow, Jonathon Shlens, and Christian Szegedy. 2015.
\newblock \href {http://arxiv.org/abs/1412.6572} {Explaining and harnessing adversarial examples}.
\newblock In \emph{International Conference on Learning Representations}.

\bibitem[{Habernal and Gurevych(2016)}]{habernal2016argument}
Ivan Habernal and Iryna Gurevych. 2016.
\newblock Which argument is more convincing? analyzing and predicting convincingness of web arguments using bidirectional lstm.
\newblock In \emph{Proceedings of the 54th Annual Meeting of the Association for Computational Linguistics (Volume 1: Long Papers)}, pages 1589--1599.

\bibitem[{Hernandez et~al.(2024)Hernandez, Sharma, Haklay, Meng, Wattenberg, Andreas, Belinkov, and Bau}]{hernandezlinearity}
Evan Hernandez, Arnab~Sen Sharma, Tal Haklay, Kevin Meng, Martin Wattenberg, Jacob Andreas, Yonatan Belinkov, and David Bau. 2024.
\newblock \href {https://openreview.net/forum?id=w7LU2s14kE} {Linearity of relation decoding in transformer language models}.
\newblock In \emph{The Twelfth International Conference on Learning Representations}.

\bibitem[{Jastrzebski et~al.(2018)Jastrzebski, Arpit, Ballas, Verma, Che, and Bengio}]{jastrzebski2018residual}
Stanis{\l}aw Jastrzebski, Devansh Arpit, Nicolas Ballas, Vikas Verma, Tong Che, and Yoshua Bengio. 2018.
\newblock Residual connections encourage iterative inference.
\newblock In \emph{International Conference on Learning Representations}.

\bibitem[{Meng et~al.(2022)Meng, Bau, Andonian, and Belinkov}]{meng2022locating}
Kevin Meng, David Bau, Alex Andonian, and Yonatan Belinkov. 2022.
\newblock Locating and editing factual associations in gpt.
\newblock \emph{Advances in Neural Information Processing Systems}, 35:17359--17372.

\bibitem[{nostalgebraist(2020)}]{lesswrong_logitlens}
nostalgebraist. 2020.
\newblock Interpreting gpt: The logit lens.
\newblock \url{https://www.lesswrong.com/posts/AcKRB8wDpdaN6v6ru/interpreting-gpt-the-logit-lens}.
\newblock Accessed: 2024-12-24.

\bibitem[{Pal et~al.(2023)Pal, Sun, Yuan, Wallace, and Bau}]{pal2023future}
Koyena Pal, Jiuding Sun, Andrew Yuan, Byron~C Wallace, and David Bau. 2023.
\newblock Future lens: Anticipating subsequent tokens from a single hidden state.
\newblock In \emph{Proceedings of the 27th Conference on Computational Natural Language Learning (CoNLL)}, pages 548--560.

\bibitem[{Sharkey et~al.(2025)Sharkey, Chughtai, Batson, Lindsey, Wu, Bushnaq, Goldowsky-Dill, Heimersheim, Ortega, Bloom et~al.}]{sharkey2025open}
Lee Sharkey, Bilal Chughtai, Joshua Batson, Jack Lindsey, Jeff Wu, Lucius Bushnaq, Nicholas Goldowsky-Dill, Stefan Heimersheim, Alejandro Ortega, Joseph Bloom, et~al. 2025.
\newblock Open problems in mechanistic interpretability.
\newblock \emph{arXiv preprint arXiv:2501.16496}.

\bibitem[{Wolf et~al.(2020)Wolf, Debut, Sanh, Chaumond, Delangue, Moi, Cistac, Rault, Louf, Funtowicz et~al.}]{wolf2020transformers}
Thomas Wolf, Lysandre Debut, Victor Sanh, Julien Chaumond, Clement Delangue, Anthony Moi, Pierric Cistac, Tim Rault, Remi Louf, Morgan Funtowicz, et~al. 2020.
\newblock Transformers: State-of-the-art natural language processing.
\newblock In \emph{Proceedings of the 2020 conference on empirical methods in natural language processing: system demonstrations}, pages 38--45.

\bibitem[{Yuksekgonul et~al.(2024)Yuksekgonul, Chandrasekaran, Jones, Gunasekar, Naik, Palangi, Kamar, and Nushi}]{yuksekgonul2024attention}
Mert Yuksekgonul, Varun Chandrasekaran, Erik Jones, Suriya Gunasekar, Ranjita Naik, Hamid Palangi, Ece Kamar, and Besmira Nushi. 2024.
\newblock \href {https://openreview.net/forum?id=gfFVATffPd} {Attention satisfies: A constraint-satisfaction lens on factual errors of language models}.
\newblock In \emph{The Twelfth International Conference on Learning Representations}.

\end{thebibliography}

\end{document}